\documentclass[11pt,a4paper]{article}
\usepackage[hyperref]{eacl2021}
\usepackage{times}
\usepackage{url}
\usepackage{latexsym}

\usepackage{amsmath}
\usepackage{amsfonts}
\usepackage{booktabs}

\usepackage{microtype}

\aclfinalcopy 

\usepackage{graphics}
\usepackage{framed,color,xcolor}
\usepackage{multirow}
\usepackage{nicefrac}
\usepackage{threeparttable}
\usepackage{todonotes}
\usepackage{multirow}
\usepackage{colortbl}

\newcommand{\UDA}{URL}
\newcommand{\CDA}{CDA}
\newcommand{\TFIDF}{$\text{TF}\!\times\!\text{IDF}$}

\newcommand{\BDAST}{WMT-16 Bilingual Document Alignment Shared Task}
\newcommand{\DSWMTTRAIN}{WMT-16 Shared Task}
\newcommand{\DSWMTDC}{WMT-16 Deep Crawl}
\newcommand{\DSCCSEVENT}{CommonCrawl Sextet}

\definecolor{pipeline-color}{rgb}{0.85,0.91,0.99}
\definecolor{saq-color}{rgb}{1.0,0.9,0.8}
\definecolor{cleanup-color}{rgb}{1.0,0.95,0.8}
\definecolor{cpdd-color}{rgb}{0.84,0.91,0.83}

\definecolor{pipeline-border-color}{rgb}{0.42,0.56,0.75}
\definecolor{saq-border-color}{rgb}{0.84,0.61,0.01}
\definecolor{cleanup-border-color}{rgb}{0.84,0.71,0.34}
\definecolor{cpdd-border-color}{rgb}{0.51,0.70,0.40}

\usepackage{soul}      
\definecolor{mygreen}{rgb}{0.0, 0.44, 0.0}

\title{\CDA: a Cost Efficient Content-based Multilingual Web Document Aligner}

\author{Thuy Vu \\
  Amazon Alexa AI\\
  Manhattan Beach, CA, USA \\
  \texttt{thuyvu@amazon.com} \\\And
  Alessandro Moschitti \\
  Amazon Alexa AI\\
  Manhattan Beach, CA, USA \\
  \texttt{amosch@amazon.com} \\}

\date{}

\begin{document}
\maketitle
\begin{abstract}
  
We introduce a Content-based Document Alignment approach ({\CDA}), an efficient method to align multilingual web documents based on content in creating parallel training data for machine translation (MT) systems operating at the industrial level. {\CDA} works in two steps: (i) projecting documents of a web domain to a shared multilingual space; then (ii) aligning them based on the similarity of their representations in such space. We leverage lexical translation models to build vector representations using {\TFIDF}. {\CDA} achieves performance comparable with state-of-the-art systems in the {\BDAST} benchmark while operating in multilingual space. Besides, we created two web-scale datasets to examine the robustness of {\CDA} in an industrial setting involving up to 28 languages and millions of documents. The experiments show that {\CDA} is robust, cost-effective, and is significantly superior in (i) processing large and noisy web data and (ii) scaling to new and low-resourced languages.

\end{abstract}

\section{Introduction}
\label{sec:intro}
Online machine translation (MT) services require industrial-scale training data, i.e., significantly large and high-quality parallel sentences, to build accurate models.
Exploiting the web for multilingual content has become a usual strategy in collecting large-scale parallel sentences for MT~\cite{uszkoreit-EtAl:2010:PAPERS,smith-EtAl:2013:ACL2013,buck-koehn:2016:WMT1}.
Structural Translation Recognition for Acquiring Natural Data (STRAND)~\cite{resnik-smith-2003-web} is a standard pipeline to extract parallel data from the web, consisting in three steps: (i) bilingual document alignment for an input set of documents, (ii) sentence alignment for each aligned document pair, and (iii) sentence filtering for non-translation or boilerplate cleaning.
The first step of identifying bilingual documents is technically challenging and made more complicated by the presence of large and noisy documents from web data.

In the~{\BDAST} (WMT16-BDAST), two standard approaches for identifying parallel pages\!\footnote{``page'' and ``document'' interchangeably refer to the content of a web page.} were studied:
1) URL matching heuristic~\cite{smith-EtAl:2013:ACL2013} as a baseline and
2) content similarity as a solution to maximize the performance in identifying parallel documents.
The benchmark on English-French document alignment task shows that the best top-1 recall (R$@1$) for each approach are $59.8$\% and $89.1$\%, respectively, as evaluated on the test set~\cite{buck-koehn:2016:WMT1,buck-koehn:2016:WMT2}.
The results, albeit conducted within English-French setting, indicate that leveraging document content can lead to a significant increase, up to $30$ percent points in recall, and contributes $\sim$50\% novel bilingual document pairs.

The URL matching heuristic approach, named {\UDA}, identifies parallel pages using language identifiers, typically from ISO 639, annotated in the addresses.
Pages, or web-documents, in different languages from a domain are aligned if their URLs are matchable after their language-identifier being removed~\cite{smith-EtAl:2013:ACL2013}.
The strategy can identify a candidate at a decent cost by comparing two URLs without significant pre-processing needed.
For example, the following URLs are a match: \texttt{xyz.ca/index.htm} and \texttt{xyz.ca/{\color{blue} fr/}index.htm} after removing ``\texttt{{\color{blue} fr/}}'' from the second URL.
On the contrary, cost is the major issue when comparing content as it often requires language-specific processing and sophisticated modelings for cross-language normalization and alignment.
The problem becomes even more challenging when dealing with web data and for low-resourced languages.

We optimize the cost for the latter approach to enable its application at scale.
Specifically, we design {\CDA} to project multilingual documents to a shared multilingual space for direct similarity measurement.
Therefore, we can run the STRAND pipeline for multiple languages at once to keep the pre-processing cost \emph{monotonic} with respect to the number of languages and documents.
In particular, we design {\CDA} with two key objectives: i) minimal data processing cost and ii) fast scaling to new languages.
The latter is also crucial to the language expansion in online MT services. Our contribution is three-fold: 

\begin{itemize}
  \item We present an optimized, efficient, and scalable framework that can perform multilingual document alignment at state-of-the-art performance in one go.
  \item To facilitate the development and evaluation, we created two web-scale datasets, which are much larger and have many more languages than what is currently publicly available.
  \item We study the contribution of {\CDA} in multiple applications involving parallel sentence extraction for MT and mutual complement between {\UDA} and {\CDA}. 
\end{itemize}

We tested {\CDA} with multiple large-scale datasets of web documents.
We also studied the applications of {\CDA} within an industrial setting, including (i) extracting more and better parallel sentences from an extremely large-scale dataset, (ii) producing better MT models, and (iii) improving yield, measured by the amount of extracted parallel content, for both {\UDA} and {\CDA} in the STRAND pipeline.
The experimental results show that, despite its minimality, {\CDA} (i) is on par with the top systems from WMT16-BDAST, which use expensive bilingual resources, and (ii) can double the amount of parallel data extracted by previous URL-based approaches.
Most importantly, we show that {\CDA} provides robust performance when dealing with millions of documents and processing up to 28 languages, including low-resourced languages. 

In the remainder of this paper, we summarize the previous work regarding document alignment in Section~\ref{sec:prework}.
We then describe the proposed system and our experiments in sections~\ref{sec:system} and \ref{sec:experiment}, respectively.
Finally, we derive the conclusions of the paper in Section~\ref{sec:conclusion}.

\section{Related Work}
\label{sec:prework}

Aligning multilingual documents is the key required processing in most multilingual text processing pipelines, including cross-lingual information retrieval~\cite{steinberger02, pouliquen-etal-2004-multilingual,VulicIvan2015Maci,jiang-etal-2020-cross} and parallel data extraction.
In the context of creating parallel training data for MT, the problem has been studied in the literature for comparable corpora~\cite{munteanu-etal-2004-improved,vu-etal-2009-feature,pal-etal-2014-automatic} and web-structured parallel data extraction~\cite{resnik-1999-mining,uszkoreit-EtAl:2010:PAPERS,buck-koehn:2016:WMT1}. We focus on the latter in this paper. 

{\BDAST} is a recent shared-task focusing on identifying bilingual documents from crawled websites~\cite{buck-koehn:2016:WMT2}.
The top 3 systems are {\bf YODA}~\cite{dara-lin-2016-yoda}, {\bf N{\sc \bf OVA}L{\sc \bf INCS}}~\cite{gomes-pereira-lopes-2016-first}, and {\bf UE{\sc \bf DIN1 COSINE}}~\cite{buck-koehn:2016:WMT2}.
The first two require costly features, such as (i) n-gram comparison after translating all non-English text into English~\cite{dara-lin-2016-yoda}, and (ii) phrase-table of statistical MT (SMT) as a dictionary~\cite{gomes-pereira-lopes-2016-first}.
{\bf UE{\sc \bf DIN1 COSINE}}~\cite{buck-koehn:2016:WMT2}, on the other hand, only uses {\TFIDF}-weighted with Cosine similarity. 
Interestingly, this method performs surprisingly well even without French-to-English translations, dropping just 3.4\% in recall, from 93.7\% to 90.3\%.
Though the finding can be due to the English and French lexicons' overlap, it suggests that {\TFIDF} with proper normalization is useful to compare document representations from sub-domains.
Our proposed method exploits this aspect. 

Given the advent of deep neural network modeling, document embeddings are among the main interests in general NLP applications~\cite{pmlr-v32-le14,cer-etal-2018-universal,el-kishky-guzman-2020-massively}.
This line of research, however, is \emph{not} technically related to our problem setting.
Specifically, the cost to run a neural inference over a web-scale setting is prohibitively high, e.g., processing a dataset of several billion pages from CommonCrawl\footnote{\url{commoncrawl.org}} is not feasible.
To have an idea, the Cloud Translation\footnote{\url{cloud.google.com/translate/pricing}} cost to translate a webpage having 20,000 characters is $\$0.4$ as of Jan 2021.

\section{Proposed System}
\label{sec:system}

We describe our method to identify parallel pages from a web domain.
Specifically, pages in different languages are projected to a shared space, where their similarity can be measured.

\paragraph{Problem Definition} Let $D=\{ d_1, d_2, \dots, d_n\}$ be the $n$ pages from a domain,
each page $d_i$ is described by its content $c_i$ in language $L_i$.
The problem is to identify all $\left(d_i, d_j\right)$ of different languages, $L_i \ne L_j$, and
$c_i$ and $c_j$ are translational equivalent.

\paragraph{Multilingual Space} Let $\mathcal{L}=\{L_1, L_2, \dots\}$ be the set of languages found in $D$ and let
    $D_i$ be the set of documents in $L_i$. Thus, $D= \bigcup_{L_i \in \mathcal{L}}^{} D_i$, where
each $D_i$ is associated with a lexicon $V_i$.
Without loss of generality, we project documents from two languages, $L_\alpha$ and $L_\beta$, into a common space as follows.
We first define alignment between two lexicons $V_\alpha$ and $V_\beta$ as:
    \begin{align*}
    \mathcal{A} & =  \big\{ \left(a, b\right): a \in V_\alpha, b \in V_\beta, \\
    		& P_{\alpha\rightarrow\beta}(a,b) + P_{\beta\rightarrow\alpha} (b,a) \\ & \ge P_{\alpha\rightarrow\beta}(a,w) + P_{\beta\rightarrow\alpha}(w,a),  \forall w \in V_\beta 
 \big\},
    \end{align*}
where $P_{\alpha\rightarrow\beta}$ and $P_{\beta\rightarrow\alpha}$ are lexical translation models from $L_\alpha$ to $L_\beta$, and vice versa\footnote{It can be easily shown that the proposed aligned is symmetric, i.e., the other condition $P_{\alpha\rightarrow\beta}(a,b) + P_{\beta\rightarrow\alpha} (b,a) \ge P_{\alpha\rightarrow\beta}(w,b) + P_{\beta\rightarrow\alpha}(b,w),  \forall w \in V_\alpha$ holds.}.
It should be noted that $\mathcal{A}$ defines a common space, $\mathbb{R}^{|\mathcal{A}|}$, where the dimensions are all word pairs. However, we can simplify the approach by mapping all languages in the space of a pivot language, i.e., $\alpha$. Thus, we define $\Pi_{\alpha}: D_{\beta} \longrightarrow \mathbb{R}^{|V_{\alpha}|}$ that maps documents $d_{\beta} \in D_{\beta}$ into the same space of $D_\alpha$, as:
    \begin{align}
        \Pi_{\alpha}(d_{\beta}) & = \Pi_{\alpha}\left(w_1, w_2, \dots, w_{|d_{\beta}|}\right) \nonumber \\ 
        & = \vec x = \left(x_1, x_2, \dots, x_{|V_\alpha|}\right) \label{map} 
    \end{align}
We define a lexical mapping for document $d_{\beta}$ as $M(d_{\beta}) = d_{\beta\rightarrow\alpha} =\{w'_1,...,w'_{|d_{\beta\rightarrow\alpha}|} | (w_i,w'_j) \in \mathcal{A}\}$, which maps $d_{\beta}$ into language $\alpha$.
Similar, we denote the mapping for a document collection $D_{\beta}$ as $D_{\beta\rightarrow\alpha} = \{M(d_{\beta}): \forall d_{\beta} \in D_{\beta} \}$. Finally, we compute the \TFIDF\ representation of $d_{\beta}$ as follows,  $\forall w_i \in V_{\alpha}$:
     \begin{itemize}
      	\item $\text{TF}(w_i)=$ number of occurrences of $w_i$ in $d_{\beta\rightarrow\alpha}$; and
      	\item $\text{IDF}(w_i)= \log\left(1 + \frac{|D_{\beta\rightarrow\alpha}|}{1 + \#(w_i,D_{\beta\rightarrow\alpha})}\right)$, where $\#\left(w, D\right)$ returns the number of documents in $D$ containing $w$.
    \end{itemize}
We compute $x_i$ in Eq.~\ref{map} using $\text{TF}(w_i)\times\text{IDF}(w_i)$.

\paragraph{Aligning Multilingual Documents} Two documents are considered a good pair if their representations are similar, according to a similarity threshold $t$.
We compute the similarity between $d_i$ and $d_j$ as the dot-product between, $v_i \cdot v_j \in [0..1]$
(we normalized the vector representations with $\ell_2$).
In practice, we use English to build the multilingual space as it is the dominant language on the Internet and in most multilingual websites.

\section{Experiments}
\label{sec:experiment}

We examine the efficacy of {\CDA} in this section.
First, we describe (i) the pipeline setup for the experiments and (ii) our effort in creating suitable benchmark data and selecting relevant resources in Section~\ref{sec:pipeline}, and Section~\ref{sec:datasetresource}, respectively.
We then address the following performance aspects of {\CDA}:

\begin{enumerate}
  \item The performance in multilingual document alignment.
  \item The impact of {\CDA}, compared to {\UDA}, in an end-to-end STRAND pipeline.
  \item The by-product applications of {\CDA} in identifying (i) novel language identifiers beyond ISO 639 for {\UDA} and (ii) web-domains containing multilingual data that are not detectable using language identifiers.
  \item The cost required to enable the support to a new language.
\end{enumerate}

\subsection{Pipeline Setup}
\label{sec:pipeline}

\begin{figure}
\centering
\includegraphics[width=1.0\linewidth]{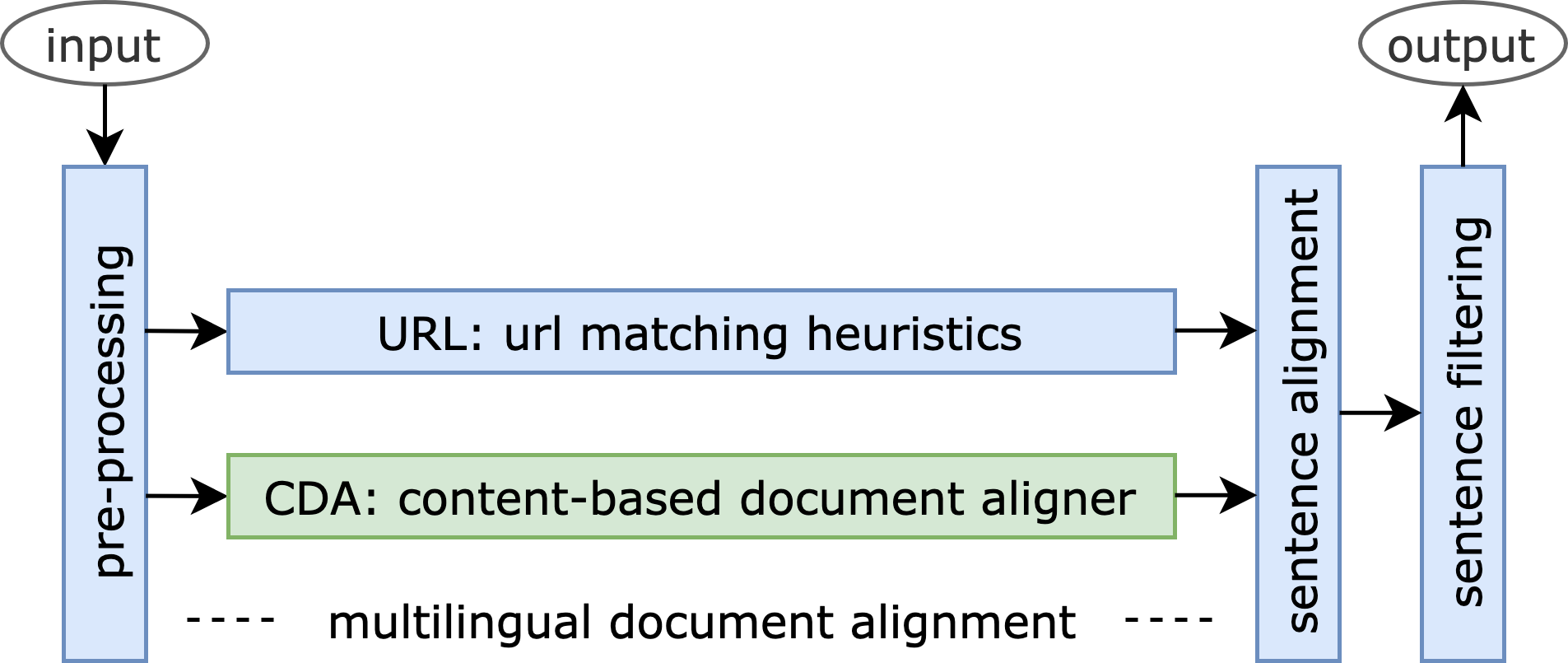}
\caption{A \fcolorbox{pipeline-border-color}{pipeline-color}{STRAND pipeline} that uses \fcolorbox{pipeline-border-color}{pipeline-color}{\UDA} to align document and \fcolorbox{cpdd-border-color}{cpdd-color}{\CDA} as a replacement.}
\label{fig:pipeline}
\end{figure}

Figure~\ref{fig:pipeline} depicts the STRAND pipeline for our experiments.

\begin{itemize}
\item The {\bf input} is constituted by web documents of multiple domains. The {\bf output} is a set of parallel sentence pairs extracted from the pipeline.
Each document has a web address and a raw HTML source.
\item The {\bf pre-processing} step groups input documents by domain to create data for document alignment step using {\UDA} and {\CDA}. For {\CDA}, additionally, it extracts the text content from HTML structure, using the following tags: \texttt{title, h1..h6, label, blockquote, dd, dt, p, pre, q, div}. This helps remove boilerplate effectively from being considered in the calculation. We use Python's \texttt{langid} package to identify the language of a page.
\item {\bf Document alignment} is performed by either {\UDA} or {\CDA}. For {\UDA}, we use a similar set of language identifiers from BDAST's baseline\footnote{\url{https://github.com/christianbuck/wmt16-document-alignment-task/blob/master/languagestripper.py}}.
\item For each aligned document pair, the {\bf sentence alignment} step aligns text segments, called sentence pair candidates, of the aligned pages based on the DOM structure~\cite{smith-EtAl:2013:ACL2013}.
\item Finally, the {\bf sentence filtering} step removes low-quality pairs~\cite{xu-koehn-2017-zipporah,prompsit:2018:WMT} or duplications.
The filter we used in this experiment has approximately 90\% F1 score for each language pair.
\end{itemize}

\subsection{Dataset and Resource}
\label{sec:datasetresource}

We describe the datasets and resources used in the experiments.

\subsubsection{Dataset}
\label{sec:dataset}

We collect and create the following datasets to study  {\CDA} performance in (i) matching parallel content, (ii) handling large datasets, and (iii) extending its use to new languages.
\paragraph{\bf \DSWMTTRAIN}
First, we use the benchmark dataset provided for WMT-16 Shared Task on Bilingual Document Alignment.
We evaluate and compare {\CDA} with other English--French document alignment methods on the BDAST's training set.
The dataset consists of 348,858 and 225,043 English and French documents from 49 web-domains, respectively.
Each document has a web address and a clean content.
Besides, French documents are translated into English using a standard SMT model.
This translation is to study the potential upper-bound performance when having full translations.
An alignment candidate has one document from each language, English or French, from the same domain. 
Thus, there are more than 4.2e9 possible alignments between the documents.
The golden data has 1,624 pairs provided by WMT16-BDAST.
In this set, the number of labeled alignments per domain ranges from 4 (e.g., \texttt{www.eohu.ca}) to 236 (e.g., \texttt{tsb.gc.ca}).
The pairs generated by a system are first filtered by 1-1 rule: each document should participate in at most one alignment.
A system is evaluated based on the recall achieved on these 1,624 pairs.

\paragraph{\bf \DSWMTDC}
The previous benchmark has two limitations.
First, the size of the dataset is relatively small compared to a typical web-scale setting\footnote{A typical multilingual domain can have thousands to millions of pages; e.g. \texttt{nato.int} and \texttt{microsoft.com} have 3e5 and 38e6 pages, respectively, indexed by Google\footnote{search results returned by Google using query ``\texttt{site:...}'' for a specific website}. }.
Second, the choice of English--French is not representative of the ultimate goal --- finding more and better parallel data to enable MT in low-resourced languages.
English--French has been the most studied pair in MT task.
Besides, their lexicons are also highly overlapped~\cite{ethnologue}.

We address this problem, creating a larger dataset of more than 14MM pages using the same set of 49 domains.
Specifically, we used these domains and URLs as seeds and recursively downloaded all reachable pages from those seeds.
We did not download pages that link to external domains.
This exercise resulted in a dataset consisting of 8.7MM and 5.5MM pages for English and 28 other languages.
These languages include: Arabic, Bulgarian, Chinese Simplified, Chinese Traditional, Croatian, Czech, Danish, Dutch, English, Estonian, Finnish, French, German, Hebrew, Hungarian, Indonesian, Italian, Japanese, Korean, Malay, Norwegian, Polish, Portuguese, Russian, Slovak, Spanish, Swedish, Thai, and Turkish.

\paragraph{\bf \DSCCSEVENT}
Previous datasets share the same domains that are heavily biased toward French content (see Table~\ref{milestone2-result-details}).
We leverage a monthly crawl from CommonCrawl, specifically \texttt{CC-MAIN-2017-17}\footnote{\url{s3://commoncrawl/crawl-data/CC-MAIN-2017-17}}, to create a better distribution dataset to validate {\CDA}.
We select pages from the crawl having pages in Chinese, Czech, Italian, Japanese, Russian, and Turkish.
We only keep the following top-level domains: \texttt{.cn}, \texttt{.tw}, \texttt{.cz}, \texttt{.it}, \texttt{.jp}, \texttt{.ru}, \texttt{.tr}, \texttt{.edu}, \texttt{.gov}, and \texttt{.org}.
The process results in a dataset of 600+ domains with 17.6MM and 4.1MM pages in English and six selected languages.
Table~\ref{table:dataset} summarizes the datasets considered in our experiments.

\begin{table}
\caption{Summary of Benchmark Datasets for Document Alignment Task. Each dataset is described by the number of web-domains (\#Dom.), English documents (\#EN-docs), non-English documents (\#XX-docs), and languages (\#$\mathcal{L}$).}
\label{table:dataset}
\centering
\resizebox{1.0\linewidth}{!}{
\begin{tabular}{rrrrl}
\toprule
{\bf Dataset}     & {\bf \#Dom.} & {\bf \#EN-docs} & {\bf \#XX-docs} & \multicolumn{1}{c}{\bf \#$\mathcal{L}$ } \\
\midrule
\DSWMTTRAIN & 49 & 348,858 & 225,043 & En+Fr \\
\DSWMTDC & 49 & 8.7MM & 5.5MM & En+28 \\
\DSCCSEVENT & 662 & 17.6MM & 4.1MM & En+6 \\
\bottomrule
\end{tabular}
}
\end{table}

\subsubsection{Resource}
\label{sec:resource}

Lexical translation dictionaries are the significant resource required in our proposed method to support a new language pair.
Our experiments used the lexical translation dictionaries created by methods introduced for traditional SMT~\cite{Brown:1993} and neural-based MT~\cite{conneau2017word}.

In particular, we use IBM-1 models for popular languages that have sufficient parallel data from general domains~\cite{koehn2005epc}.
We use the GIZA++ toolkit~\cite{och03:asc} to create IBM-1 models.
Collecting such parallel data for the low-resourced languages is generally challenging. 
We instead leverage the advances in multilingual embeddings from the MUSE project\footnote{github.com/facebookresearch/MUSE}~\cite{conneau2017word}.
We create translation probability between two words by their normalized embedding similarity score.

\subsection{Bilingual Document Alignment Results}
\label{sec:wmt16}

We evaluate the performance of {\CDA} under the {\DSWMTTRAIN} benchmark.
We conduct experiments on both settings, using the original text and using full translations.
The latter setting allows us to understand the possible benefit of the expensive step, full document translation.
Besides, the construction of $V_{L_i}$ is crucial to the distinction of the representations.
Therefore, we examine the impact of the vocabulary size of $V_{L_i}$ to the alignment result.
Specifically, we construct $V_{L_i}$ by selecting the top frequent tokens after removing stop-words and the first $k$ frequent tokens. We empirically set $k$ to be 100.
We experiment with three different sizes for $V_{L_i}$: 2,000, 10,000, and 20,000.
Finally, we compare the results of {\CDA} with the baseline {\UDA} and the top-3 systems of the {\BDAST}.
The evaluation metric is the percentage of the 1,624 golden pairs found in the top-1 alignment for each English document.
Table~\ref{table:bdast} shows the result.

\begin{table}
\caption{Comparison of the baseline {\UDA}, top-3 performing systems from WMT16-BDAST and {\CDA} under different settings of vocabulary size, using either original text (original documents in English and French) or translated text (documents in English and the English translations of documents in French)}
\label{table:bdast}
\centering
\begin{threeparttable}
\resizebox{1.0\linewidth}{!}{
\begin{tabular}{rcc}
\toprule
\multicolumn{1}{c}{baseline: {\UDA}} & \multicolumn{2}{c}{67.9} \\
\midrule
\shortstack[c]{\bf content-based\\ \bf systems} & \multicolumn{1}{l}{\shortstack[c]{\bf align w.\\ \bf org. text}} & \multicolumn{1}{l}{\shortstack[c]{\bf align w.\\ \bf trans. text}}     \\
\midrule
YODA & \multirow{2}{*}{n/a} & \multirow{2}{*}{\bf 93.71} \\
~\cite{dara-lin-2016-yoda}  &                    \\
N{\sc OVA}L{\sc INCS} & \multirow{2}{*}{90.50} & \multirow{2}{*}{n/a} \\
~\cite{gomes-pereira-lopes-2016-first}  &                    \\
UE{\sc DIN1 COSINE} & \multirow{2}{*}{90.25\tnote{$\dagger$}} & \multirow{2}{*}{93.65\tnote{$\dagger$}} \\
~\cite{buck-koehn:2016:WMT2}  &                    \\
\midrule
CDA: \textbar{}V\textbar{}=2,000  & 87.17 & 90.95 \\
CDA: \textbar{}V\textbar{}=10,000 & {\bf 90.40} & 91.02 \\
CDA: \textbar{}V\textbar{}=20,000 & 90.33 & 89.90 \\
\bottomrule
\end{tabular}
}
\begin{tablenotes}
  \item \small{[$\dagger$] \emph{average performance of valid and test splits}}
\end{tablenotes}  
\end{threeparttable}
\end{table}

For alignments using original text, the results indicate that {\CDA} achieves similar performance with the state-of-the-art methods from BDAST.
The result shows the efficacy of the proposed alignment method.
The results also show that the vocabulary size, or the vector representations' size, impacts the performance.
$|V|=10,000$ yields the best result among the three settings.
Second, even though using full translation is better, the performance gains are negligible with respect to the processing cost required for building the MT models and translating all the data into an anchor language.
Since {\CDA} does not exploit bi-gram features, its performance is relatively lower, up to 3\%, compared to the state of the art.
In short, the result suggests an optimal configuration for {\CDA} with a vocabulary size of 10,000.

\subsection{Multilingual Document Alignment Results}
\label{sec:wmt16}

It was showed in WMT16-BDAST that content-based methods could add 60\% more English--French document pairs compared to {\UDA}.
This section aims to verify this in a multilingual setting, mainly when operating with a significantly higher number of languages and domains  using {\DSWMTDC} and {\DSCCSEVENT}, respectively.

\paragraph{\bf On \DSWMTDC} Table~\ref{milestone2-result-details} shows the number of parallel documents and sentences extracted by an end-to-end STRAND pipeline, after filtering and duplication removal, described in Section~\ref{sec:pipeline}.
The result shows that {\CDA} contributes an extra of $53\%$, $75\%$, and $195\%$ in \emph{clean} parallel sentences compared to {\UDA} for French alone, and when French is and is not considered, in this more extensive and more realistic setting, respectively.
It also suggests that {\CDA} is effective and can significantly increase the number of parallel sentences extracted for low-resourced languages.
Finally, the result indicates that our proposed method is robust in a multilingual setting.

\begin{table}[t]
\centering
\caption{Comparison of yields, number of clean document pairs and sentence pairs, produced via {\UDA} and {\CDA} on {\DSWMTDC}. Column {\bf \CDA \textbackslash{} \UDA} reports the number of novel sentence pairs exclusively extracted via {\CDA}.}
\label{milestone2-result-details}
\resizebox{1\linewidth}{!}{
\begin{tabular}{r|rr|rrr}
\toprule
\multirow{2}{*}{\bf Language}  & \multicolumn{2}{c|}{\bf \# Document Pairs} & \multicolumn{3}{c}{\bf \# Sentence Pairs}                     \\
                            & \multicolumn{1}{c}{\bf \UDA} & \multicolumn{1}{c|}{\bf \CDA} & \multicolumn{1}{c}{\bf \UDA} & \multicolumn{1}{c}{\bf \CDA} & \multicolumn{1}{c}{\bf \CDA\!\textbackslash{}\!\UDA} \\
\midrule
Arabic                      & ~~~~~ 3,266   & ~~~~~ 2,896   & ~~~~~~ 36,262    & ~~~~~~ 39,590   & ~~~~~~ 24,065     \\
Bulgarian                   & ~~~~~ 1,184   & ~~~~~ 1,070   & ~~~~~~~~ 9,292   & ~~~~~~~~ 1,748  & ~~~~~~~~ 1,359    \\
Chinese-S                   & ~~~~~ 2,805   & ~~~~~ 2,160   & ~~~~~~ 30,519    & ~~~~~~ 27,666   & ~~~~~~ 13,289     \\
Chinese-T                   & ~~~~~~~~ 316  & ~~~~~~~~ 102  & ~~~~~~~~ 2,584   & ~~~~~~~~ 2,055  & ~~~~~~~~~~~ 374   \\
Croatian                    & ~~~~~~~~ 704  & ~~~~~ 3,119   & ~~~~~~~~~~~ 889  & ~~~~~~ 56,300   & ~~~~~~ 55,854     \\
Czech                       & ~~~~~~~~~~ 29 & ~~~~~~~~ 241  & ~~~~~~~~~~~~~ 77 & ~~~~~~~~ 7,264  & ~~~~~~~~ 7,248    \\
Danish                      & ~~~~~~~~ 137  & ~~~~~ 2,932   & ~~~~~~~~~~~ 693  & ~~~~~~ 39,488   & ~~~~~~ 38,996     \\
Deutsch                     & ~~~~~ 5,525   & ~~~~~ 8,863   & ~~~~~~ 83,663    & ~~~~ 170,932    & ~~~~ 113,851      \\
Dutch                       & ~~~~~~~~ 599  & ~~~~~ 2,407   & ~~~~~~~~ 8,228   & ~~~~~~ 79,293   & ~~~~~~ 79,146     \\
Farsi                       & ~~~~~ 1,316   & ~~~~~ 1,404   & ~~~~~~ 14,697    & ~~~~~~ 13,875   & ~~~~~~~~ 6,122    \\
Finnish                     & ~~~~~~~~ 170  & ~~~~~ 1,313   & ~~~~~~~~~~~ 355  & ~~~~~~ 12,403   & ~~~~~~ 12,229     \\
French                      & ~ 115,671     & ~ 143,972     & ~ 2,653K      & ~ 3,568K     & ~ 1,411K       \\
Hebrew                      & ~~~~~~~~ 209  & ~~~~~~~~ 140  & ~~~~~~~~ 7,742   & ~~~~~~~~ 5,295  & ~~~~~~~~~~~~~ 83  \\
Hungarian                   & ~~~~~ 1,253   & ~~~~~ 1,382   & ~~~~~~ 10,494    & ~~~~~~~~ 6,158  & ~~~~~~~~ 4,448    \\
Indonesian                  & ~~~~~~~~ 368  & ~~~~~~~~ 551  & ~~~~~~~~~~~ 625  & ~~~~~~~~ 1,204  & ~~~~~~~~~~~ 900   \\
Italian                     & ~~~~~ 6,644   & ~~~~~ 7,310   & ~~~~~~ 57,977    & ~~~~~~ 94,098   & ~~~~~~ 55,802     \\
Japanese                    & ~~~~~~~~ 823  & ~~~~~ 1,475   & ~~~~~~~~ 6,593   & ~~~~~~ 14,720   & ~~~~~~ 11,138     \\
Korean                      & ~~~~~~~~ 913  & ~~~~~~~~ 136  & ~~~~~~ 13,365    & ~~~~~~~~ 2,229  & ~~~~~~~~~~~ 224   \\
Malay                       & ~~~~~ 1,040   & ~~~~~ 1,904   & ~~~~~~~~ 8,467   & ~~~~~~ 13,088   & ~~~~~~~~ 7,213    \\
Norwegian                   & ~~~~~~~~~~ 67 & ~~~~~ 1,875   & ~~~~~~~~~~~ 196  & ~~~~~~ 35,362   & ~~~~~~ 35,273     \\
Polish                      & ~~~~~~~~ 557  & ~~~~~~~~ 934  & ~~~~~~~~ 9,685   & ~~~~~~ 21,528   & ~~~~~~ 17,255     \\
Portugese                   & ~~~~~ 1,545   & ~~~~~ 6,200   & ~~~~~~ 12,294    & ~~~~ 104,561    & ~~~~~~ 96,850     \\
Russian                     & ~~~~~ 3,984   & ~~~~~ 2,475   & ~~~~~~ 36,565    & ~~~~~~ 55,010   & ~~~~~~ 40,722     \\
Slovak                      & ~~~~~~~~ 170  & ~~~~~~~~ 850  & ~~~~~~~~~~~ 211  & ~~~~~~~~ 2,157  & ~~~~~~~~ 2,106    \\
Spanish                     & ~~~~~ 8,334   & ~~~ 21,765    & ~~~~ 114,874     & ~~~~ 317,430    & ~~~~ 252,523      \\
Swedish                     & ~~~~~~~~~~ 83 & ~~~~~ 2,394   & ~~~~~~~~ 2,773   & ~~~~~~ 49,420   & ~~~~~~ 49,238     \\
Thai                        & ~~~~~~~~~~ 82 & ~~~~~~~~~~ 10 & ~~~~~~~~~~~ 830  & ~~~~~~~~~~~ 259 & ~~~~~~~~~~~~~ 40  \\
Turkish                     & ~~~~~ 1,057   & ~~~~~ 2,041   & ~~~~~~~~ 9,598   & ~~~~~~ 18,412   & ~~~~~~ 10,789     \\
\midrule
All                         & ~ 159,343     & ~ 222,283     & ~ 3,134K      & ~ 4,761K     & ~ 2,349K       \\
All\textbackslash{}French & ~~~ 43,672    & ~~~ 78,311    & ~~~~ 481,333     & ~ 1,193K     & ~~~~ 937,683      \\
\bottomrule
\end{tabular} 
}
\end{table}

\paragraph{\bf On \DSCCSEVENT} Table~\ref{table:commoncrawl} shows the result in English parallel tokens extracted from the pipeline using {\UDA} and {\CDA} in the document alignment step.
The result shows similar gains as in the previous experiment, except for Czech --- increasing 7\!$\times$ more parallel tokens.
Our post-hoc analysis discovers non-standard language identifiers missing for {\UDA} processing, e.g., \texttt{ces} or \texttt{cesky}.

To confirm the study, we randomly selected 1,320 English--Turkish document pairs identified by {\CDA} for human verification since we do not have annotated data.
The outcome indicates that the accuracy of the document pairs is at 91.5\%.
Information on these datasets is described here: \url{github.com/alexa/wqa\_dataset}.

\begin{table}[t]
\centering
\caption{Parallel English tokens extracted by {\CDA} and {\UDA} on {\DSCCSEVENT}}
\label{table:commoncrawl}
\resizebox{1\linewidth}{!}{
\begin{tabular}{lrrrr}
\toprule
          {\bf Lang.}& \multicolumn{1}{l}{\bf \#Dom.} & \multicolumn{1}{l}{\bf \#EN-docs} & \multicolumn{1}{l}{\bf \#XX-docs} & \multicolumn{1}{l}{\bf $\frac{|{\bf \CDA}|}{|{\bf \UDA}|}$}  \\
\midrule
Turkish  & 37                          & 1,434,923                   & 71,034                      & 1.77                           \\
Czech    & 69                          & 2,333,914                   & 831,072                     & 7.07                           \\
Japanese & 84                          & 2,097,664                   & 757,872                     & 1.90                           \\
Russian  & 125                         & 2,918,594                   & 1,163,258                   & 1.09                           \\
Italian  & 239                         & 5,770,684                   & 1,112,868                   & 1.05                           \\
Chinese  & 108                         & 3,061,782                   & 207,211                     & 0.99                           \\
\midrule
All      & 662                         & 17,617,561                    & 4,143,315                     & 1.24                          \\
\bottomrule
\end{tabular}
}
\end{table}

\subsection{Industrial Benchmarks}
We conducted multiple internal experiments to examine the performance of {\CDA} over {\UDA} under an industrial setting.
Specifically, we focus on three application aspects of {\CDA}: (i) robustness,  (ii) identifying non-standard language identifiers for {\UDA},  (iii) identifying multilingual web-domains. 
Due to business security reasons, we do not name the specific languages considered in this study.
We do not provide some details of the experiment setting, which are not critical to illustrate our findings.

\subsubsection{Robustness Benchmark}
We ran the STRAND pipeline end-to-end to extract parallel sentence pairs from document pairs identified by {\UDA} and {\CDA} replacing {\UDA}.
We employ a crawl dataset larger than a typical monthly crawl archive from CommonCrawl.
The dataset is also considered densely multilingual.
We target six mid-tier languages that are \emph{not} in the top-10 high-resourced languages.
It shows that {\CDA} can increase additional 27\% English parallel tokens over the selected languages.

\begin{figure}
\centering
\includegraphics[width=1.0\linewidth]{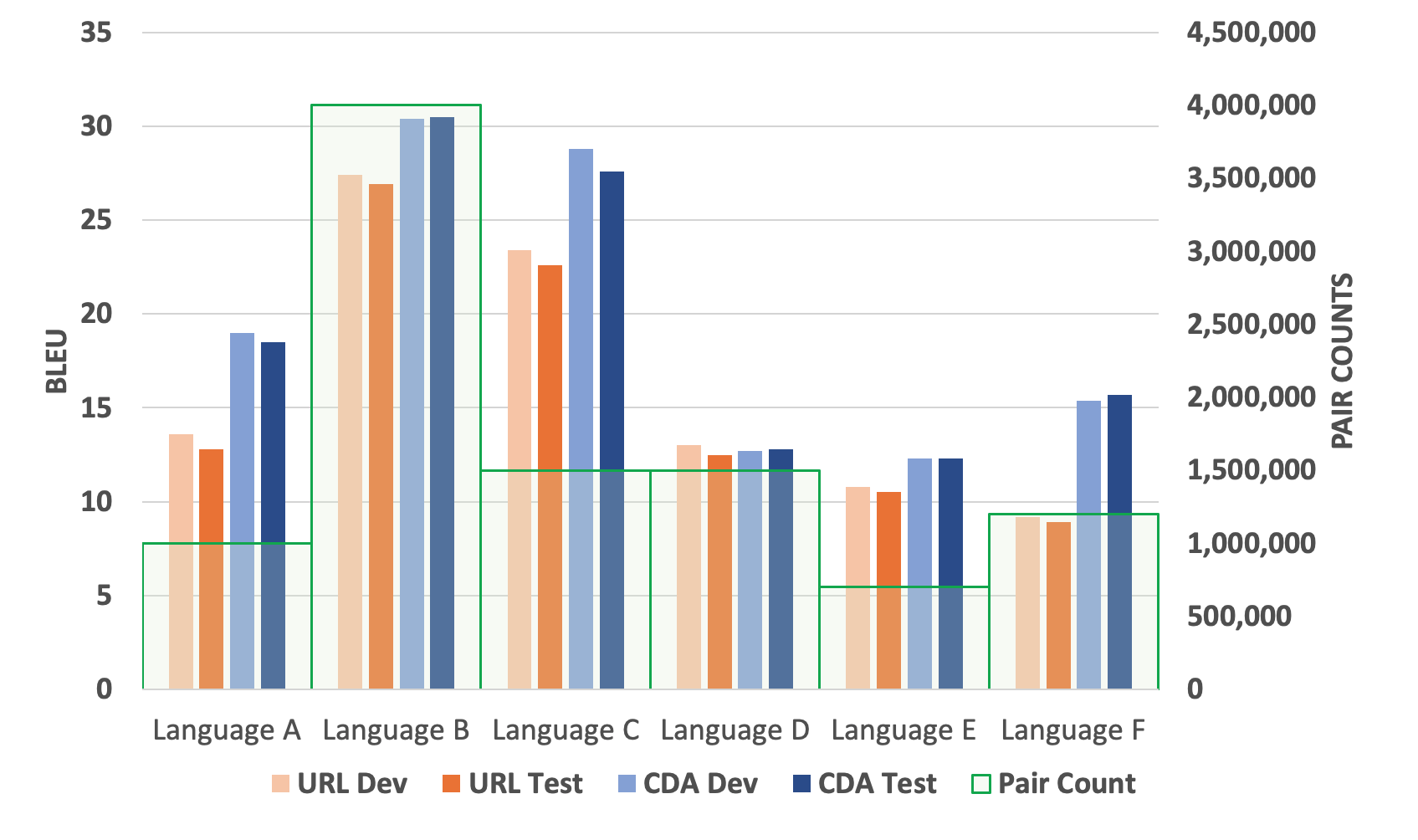}
\caption{Automatic evaluation of MT models trained by the same amount of pairs sampled from pairs extracted exclusively by either {\UDA} or {\CDA}. The primary y-axis (left) indicates the BLEU score, while the secondary y-axis (right) indicates the number of parallel pairs.}
\label{fig:compareinternal}
\end{figure}

\paragraph{Automatic Evaluation} We first study the quality of the extracted parallel data, especially the addition of 27\% produced by {\CDA}, using automatic MT evaluation.
Specifically, for each language, we compare the translation models trained by two equal-sized parallel sentence pairs sampled from the exclusive pairs extracted by {\UDA} and {\CDA} individually, i.e., after removing common pairs extracted by both methods.
We train vanilla \texttt{seq2seq} models using Sockeye\!\footnote{https://github.com/awslabs/sockeye}.
We report the MT performance in BLEU scores on our MT evaluation data in Figure~\ref{fig:compareinternal}. 
The results indicate that the models trained using novel sentence pairs extracted by {\CDA} consistently give better translation models. 

\paragraph{Human Evaluation} We had linguists manually verify the extracted parallel sentence pairs produced by the pipeline using either {\UDA} or {\CDA}.
Specifically, we randomly sampled 500 sentence pairs extracted from each pipeline using either {\UDA} and {\CDA} for Language A and Language E for human evaluation (we do not remove the common pairs in this evaluation).
The selection of these languages is based on their low performance reported during the automatic evaluation in Figure~\ref{fig:compareinternal}.
Table~\ref{human-result} shows the result in terms of precision and recall.
In general, we find that the quality of pairs produced by both methods is typically comparable.
The result also confirms the robustness of our proposed {\CDA} under stress evaluations.

\begin{table}
\caption{Human verification for parallel sentence pairs extracted from {\UDA} and {\CDA} in precision (P) and recall (R).}
\label{human-result}
\centering
\resizebox{.8\linewidth}{!}{
\begin{tabular}{rcc}
\toprule
    & {\bf \UDA} & {\bf \CDA} \\
\midrule
Language A & P=96, R=87 & P=98, R=89 \\
Language E & P=94 ,R=81 & P=92, R=90 \\
\bottomrule
\end{tabular}
}
\end{table}

\subsubsection{Identifying Non-standard Language Identifiers for {\UDA}}
Even though {\UDA} method operates decently fast, it requires language identifiers usually collected manually.
This task is challenging as language identifiers used in different web-domains typically do not follow any standard.
For example, the language patterns for Czech may include \verb+czech+, \verb+cze+, \verb+ces+, \verb+cz+, \verb+cs+, \verb+cesky+.
The result in Table~\ref{table:commoncrawl} also suggests the limitation of {\UDA} for Czech language.
In this exercise, we examine URL pairs matched by {\CDA} method to extract relevant language identifiers.
Specifically, we focus on URL pairs distanced by \emph{one} token and extract the different tokens as candidates.
For example, \verb+en+ and \verb+vi_vn+ are extracted as candidates for
    (\texttt{www.visitsingapore.com/{\color{red}en}/},
    \texttt{www.visitsingapore.com/{\color{red}vi\_vn}/}).
We curated the candidates, identified additional novel language identifiers, and feeded them to {\UDA} method.
The exercise helped significantly increase the yields for multiple low-resourced languages at 453\%, 295\%, and 266\%.
For example, we found additional identifiers for Chinese language:
\texttt{chs}, \texttt{chn}, \texttt{c}, \texttt{zho}, \texttt{zht}, \texttt{cht}, \texttt{webcn}, \texttt{sc}, \texttt{tc}, \texttt{chinese\_gb}, \texttt{chinese\_big5}, besides other popular \texttt{zh}, \texttt{chi}, and \texttt{zho}.

\subsubsection{Identifying Multilingual Web-domains}
We study the application of {\UDA} and {\CDA} in identifying densely multilingual web-domains.
Specifically, we compare the yield of parallel content, in the total number of extracted parallel English tokens, from two different datasets processed by the same pipeline.
The datasets differ in whether their web-domains are identified as multilingual by {\UDA} or {\CDA}.

On a sufficiently large dataset, we first ran {\UDA} and selected those web-domains having at least 100 candidate pairs.
Subsequently, we ran {\CDA} and selected those with at least 100 candidate pairs on the remaining of the dataset, i.e., those not selected by {\UDA}.
We randomly selected 10,000 domains from each group to create the two datasets, namely ``Domains by {\UDA}'' and ``Domains by {\CDA},'' respectively.
We applied the same pipeline using both methods for aligning documents on each dataset.
We then computed the yield of parallel English tokens extracted from each setting.
Table~\ref{result-7t} shows the results.
\begin{table}[t]
\caption{Number of parallel English tokens (in MM) from two deep crawls seeded by {\UDA} and {\CDA}.}
\label{result-7t}
\centering
\resizebox{1\linewidth}{!}{
\begin{tabular}{rrr}
\toprule
& \shortstack[c]{\bf Domains by\\ \bf {\UDA}} & \shortstack[c]{\bf Domains by\\ \bf {\CDA}} \\
\midrule
Parallel Token Count &  23.3MM &  69.5MM \\
Host Count &  10,000 & 10,000 \\
Dataset Size (TB) &  6 &  3.1 \\
\bottomrule
\end{tabular}}
\end{table}
These indicate that {\CDA} can identify densely multilingual web-domains effectively.

In particular, given the same number of web-domains, the dataset identified by {\CDA} can produce almost 3\!$\times$ more parallel data with a size of only half of the dataset identified by {\UDA}.
This finding suggests that the yield of parallel content from web-domains identified by {\CDA} is 6\!$\times$ higher than those identified by {\UDA}.
The finding is essential in optimizing the parallel extraction pipeline and identifying better densely multilingual web content.

\subsection{Cost Analysis}
\label{sec:cost}
As presented, we anticipate two cost types when extending {\CDA} to support a new language: (i) building a lexical translation model and (ii) processing more documents.
The former is a one-time cost, while the latter is dataset dependent.

Specifically, we have shown in Section~\ref{sec:resource} that a lexical translation model can be built using either statistical method IBM-1 with parallel data or neural-based unsupervised method~\cite{conneau2017word}; we observed comparable performance of {\CDA} when using a model built by these methods.
Given the rapid advance in deep neural language models, it is increasingly possible to obtain such resources for low-resourced languages.
This suggests that we will be able to leverage recent advances in neural-based NLP to continuously extend {\CDA} for many more languages. 

Regarding the execution time, the primary bottleneck typically is due to the scoring of all possible alignments between English and non-English documents.
Even though this scoring step is quadratic, this workload is perfectly parallel.
With proper engineering optimization, we empirically found out that it is possible to bring the run-time for {\CDA} to be within 2.5\!$\times$ than the one of {\UDA}'s for 20 low-resourced languages and on a sufficiently large dataset.
This optimized cost is crucial in enabling a spectrum of multilingual applications, including cross-lingual information retrieval and enabling MT services for scarce languages.

\section{Conclusion}
\label{sec:conclusion}

We presented our content-based document alignment for web data, {\CDA}, which projects multilingual documents to a common space for similarity calculation.
We also described our effort to collect and create benchmark datasets to study different performance aspects of the proposed method.
The results show that {\CDA} is efficient when projecting multilingual documents in one go for as many as 28 languages.
Moreover, we also explain the different types of benchmarking for {\CDA} under industrial settings.

The results show that our proposed method is robust when processing huge datasets and useful in identifying non-standard language identifiers and multilingual web-domains.
Finally, and most importantly, the only significant resource required by {\CDA} is the lexical translation dictionary: this can be easily built thanks to the recent advance in learning multilingual embeddings.

Future applications of {\CDA} can be many.
For example, the URLs paired with {\CDA} can be used to improve the coverage for {\UDA}-based methods (e.g., Czech case in Table~\ref{table:commoncrawl}) and to study the web structure of multilingual content. 
Moreover, the robustness and extensibility of {\CDA} make it applicable to other multilingual processing systems, including cross-lingual search and retrieval.

\section*{Acknowledgements}
We thank Chris Bissell, Jeremiah Hankins, Kwang Hyun Jang, Daniel Marcu, Masud Moshtaghi, Akash Patel, Chris de Vries, Renxia Wang, William Wong, and members of the Alexa AI Search team in Manhattan Beach, CA for their helpful feedback on earlier drafts of the manuscript.
We also thank the anonymous reviewers for their thoughtful suggestions which led to an improved manuscript.

\bibliography{cda}

\begin{thebibliography}{25}
\expandafter\ifx\csname natexlab\endcsname\relax\def\natexlab#1{#1}\fi

\bibitem[{Brown et~al.(1993)Brown, Della-Pietra, Della-Pietra, and
  Mercer}]{Brown:1993}
Peter~F. Brown, Stephen Della-Pietra, Vincent Della-Pietra, and Robert~L.
  Mercer. 1993.
\newblock The mathematics of statistical machine translation.
\newblock \emph{Comp.Ling.}

\bibitem[{Buck and Koehn(2016{\natexlab{a}})}]{buck-koehn:2016:WMT1}
Christian Buck and Philipp Koehn. 2016{\natexlab{a}}.
\newblock Findings of the wmt 2016 bilingual document alignment shared task.
\newblock In \emph{WMT 2016}, pages 554--563, Berlin, Germany.

\bibitem[{Buck and Koehn(2016{\natexlab{b}})}]{buck-koehn:2016:WMT2}
Christian Buck and Philipp Koehn. 2016{\natexlab{b}}.
\newblock {Quick and Reliable Document Alignment via TF/IDF-weighted Cosine
  Distance}.
\newblock In \emph{WMT 2016}, pages 672--678, Berlin, Germany.

\bibitem[{Cer et~al.(2018)Cer, Yang, Kong, Hua, Limtiaco, St.~John, Constant,
  Guajardo-Cespedes, Yuan, Tar, Strope, and Kurzweil}]{cer-etal-2018-universal}
Daniel Cer, Yinfei Yang, Sheng-yi Kong, Nan Hua, Nicole Limtiaco, Rhomni
  St.~John, Noah Constant, Mario Guajardo-Cespedes, Steve Yuan, Chris Tar,
  Brian Strope, and Ray Kurzweil. 2018.
\newblock \href {https://doi.org/10.18653/v1/D18-2029} {Universal sentence
  encoder for {E}nglish}.
\newblock In \emph{EMNLP: System Demonstrations}, pages 169--174, Brussels,
  Belgium.

\bibitem[{Dara and Lin(2016)}]{dara-lin-2016-yoda}
Aswarth~Abhilash Dara and Yiu-Chang Lin. 2016.
\newblock {YODA} system for {WMT}16 shared task: Bilingual document alignment.
\newblock In \emph{WMT 2016}, Berlin, Germany.

\bibitem[{El-Kishky and Guzm{\'a}n(2020)}]{el-kishky-guzman-2020-massively}
Ahmed El-Kishky and Francisco Guzm{\'a}n. 2020.
\newblock Massively multilingual document alignment with cross-lingual
  sentence-mover{'}s distance.
\newblock In \emph{AACL-IJCNLP 2020}, pages 616--625, Suzhou, China.

\bibitem[{Gomes and Pereira~Lopes(2016)}]{gomes-pereira-lopes-2016-first}
Lu{\'\i}s Gomes and Gabriel Pereira~Lopes. 2016.
\newblock First steps towards coverage-based document alignment.
\newblock In \emph{WMT 2016}, pages 697--702, Berlin, Germany.

\bibitem[{Jiang et~al.(2020)Jiang, El-Jaroudi, Hartmann, Karakos, and
  Zhao}]{jiang-etal-2020-cross}
Zhuolin Jiang, Amro El-Jaroudi, William Hartmann, Damianos Karakos, and Lingjun
  Zhao. 2020.
\newblock \href {https://www.aclweb.org/anthology/2020.clssts-1.5}
  {Cross-lingual information retrieval with {BERT}}.
\newblock In \emph{Proceedings of the workshop on Cross-Language Search and
  Summarization of Text and Speech (CLSSTS2020)}, pages 26--31, Marseille,
  France. European Language Resources Association.

\bibitem[{Koehn(2005)}]{koehn2005epc}
Philipp Koehn. 2005.
\newblock {Europarl: A Parallel Corpus for Statistical Machine Translation}.
\newblock In \emph{MT Summit 2005}, pages 79--86, Phuket, Thailand.

\bibitem[{Lample et~al.(2018)Lample, Conneau, Ranzato, Denoyer, and
  Jégou}]{conneau2017word}
Guillaume Lample, Alexis Conneau, Marc'Aurelio Ranzato, Ludovic Denoyer, and
  Hervé Jégou. 2018.
\newblock \href {https://openreview.net/forum?id=H196sainb} {Word translation
  without parallel data}.
\newblock In \emph{ICLR 2018}.

\bibitem[{Le and Mikolov(2014)}]{pmlr-v32-le14}
Quoc Le and Tomas Mikolov. 2014.
\newblock \href {http://proceedings.mlr.press/v32/le14.html} {Distributed
  representations of sentences and documents}.
\newblock volume~32 of \emph{Proceedings of Machine Learning Research}, pages
  1188--1196, Bejing, China. PMLR.

\bibitem[{Lewis(2009)}]{ethnologue}
M.~Paul Lewis, editor. 2009.
\newblock \emph{Ethnologue: Languages of the World}, sixteenth edition.
\newblock SIL International, Dallas, TX, USA.

\bibitem[{Munteanu et~al.(2004)Munteanu, Fraser, and
  Marcu}]{munteanu-etal-2004-improved}
Dragos~Stefan Munteanu, Alexander Fraser, and Daniel Marcu. 2004.
\newblock Improved machine translation performance via parallel sentence
  extraction from comparable corpora.
\newblock In \emph{{HLT}-{NAACL} 2004}, pages 265--272, Boston, Massachusetts,
  USA.

\bibitem[{Och and Ney(2003)}]{och03:asc}
Franz~Josef Och and Hermann Ney. 2003.
\newblock A systematic comparison of various statistical alignment models.
\newblock \emph{Computational Linguistics}, 29(1):19--51.

\bibitem[{Pal et~al.(2014)Pal, Pakray, and Naskar}]{pal-etal-2014-automatic}
Santanu Pal, Partha Pakray, and Sudip~Kumar Naskar. 2014.
\newblock Automatic building and using parallel resources for {SMT} from
  comparable corpora.
\newblock In \emph{{H}y{T}ra}.

\bibitem[{Pouliquen et~al.(2004)Pouliquen, Steinberger, Ignat, K{\"a}sper, and
  Temnikova}]{pouliquen-etal-2004-multilingual}
Bruno Pouliquen, Ralf Steinberger, Camelia Ignat, Emilia K{\"a}sper, and Irina
  Temnikova. 2004.
\newblock \href {https://www.aclweb.org/anthology/C04-1138} {Multilingual and
  cross-lingual news topic tracking}.
\newblock In \emph{{COLING} 2004: Proceedings of the 20th International
  Conference on Computational Linguistics}, pages 959--965, Geneva,
  Switzerland. COLING.

\bibitem[{Resnik(1999)}]{resnik-1999-mining}
Philip Resnik. 1999.
\newblock Mining the web for bilingual text.
\newblock In \emph{ACL 1999}, pages 527--534, College Park, Maryland, USA.

\bibitem[{Resnik and Smith(2003)}]{resnik-smith-2003-web}
Philip Resnik and Noah~A. Smith. 2003.
\newblock The web as a parallel corpus.
\newblock \emph{Computational Linguistics}, 29(3):349--380.

\bibitem[{S\'{a}nchez-Cartagena et~al.(2018)S\'{a}nchez-Cartagena,
  Ba{\~n}\'{o}n, Ortiz-Rojas, and Ram\'{i}rez-S\'{a}nchez}]{prompsit:2018:WMT}
V\'{i}ctor~M. S\'{a}nchez-Cartagena, Marta Ba{\~n}\'{o}n, Sergio Ortiz-Rojas,
  and Gema Ram\'{i}rez-S\'{a}nchez. 2018.
\newblock Prompsit's submission to wmt 2018 parallel corpus filtering shared
  task.
\newblock In \emph{WMT 2018}, Brussels, Belgium.

\bibitem[{Smith et~al.(2013)Smith, Saint-Amand, Plamada, Koehn, Callison-Burch,
  and Lopez}]{smith-EtAl:2013:ACL2013}
Jason~R. Smith, Herve Saint-Amand, Magdalena Plamada, Philipp Koehn, Chris
  Callison-Burch, and Adam Lopez. 2013.
\newblock Dirt cheap web-scale parallel text from the common crawl.
\newblock In \emph{ACL 2013}, pages 1374--1383, Sofia, Bulgaria.

\bibitem[{Steinberger et~al.(2002)Steinberger, Pouliquen, and
  Hagman}]{steinberger02}
Ralf Steinberger, Bruno Pouliquen, and Johan Hagman. 2002.
\newblock Cross-lingual document similarity calculation using the multilingual
  thesaurus {EUROVOC}.
\newblock In \emph{Computational Linguistics and Intelligent Text Processing,
  Third International Conference, CICLing'2002}, number 2276 in Lecture Notes
  in Computer Science, LNCS, pages 415--424. Springer-Verlag.

\bibitem[{Uszkoreit et~al.(2010)Uszkoreit, Ponte, Popat, and
  Dubiner}]{uszkoreit-EtAl:2010:PAPERS}
Jakob Uszkoreit, Jay Ponte, Ashok Popat, and Moshe Dubiner. 2010.
\newblock Large scale parallel document mining for machine translation.
\newblock In \emph{COLING 2010}, China.

\bibitem[{Vu et~al.(2009)Vu, Aw, and Zhang}]{vu-etal-2009-feature}
Thuy Vu, AiTi Aw, and Min Zhang. 2009.
\newblock Feature-based method for document alignment in comparable news
  corpora.
\newblock In \emph{{EACL} 2009}, pages 843--851, Athens, Greece.

\bibitem[{Vulic and Moens(2015)}]{VulicIvan2015Maci}
Ivan Vulic and Marie-Francine Moens. 2015.
\newblock \href
  {$$Uhttps://lirias.kuleuven.be/retrieve/322606$$DVulicMoensSIGIR2015Final.pdf
  [Available for KU Leuven users]} {Monolingual and cross-lingual information
  retrieval models based on (bilingual) word embeddings}.
\newblock pages 363--372. ACM; New York, NY.

\bibitem[{Xu and Koehn(2017)}]{xu-koehn-2017-zipporah}
Hainan Xu and Philipp Koehn. 2017.
\newblock {Z}ipporah: a fast and scalable data cleaning system for noisy
  web-crawled parallel corpora.
\newblock In \emph{EMNLP 2017}, Denmark.

\end{thebibliography}
\bibliographystyle{acl_natbib}

\end{document}